\pdfoutput=1
\documentclass[11pt]{article}

\usepackage{EMNLP2022}

\usepackage{times}
\usepackage{latexsym}
\usepackage{array}
\usepackage{pdfpages}
\usepackage{times}
\usepackage{latexsym}
\usepackage{amsmath}
\usepackage{graphicx}
\usepackage{textcomp}
\usepackage[shortlabels]{enumitem}

\usepackage[T1]{fontenc}
\usepackage[utf8]{inputenc}

\usepackage{microtype}
\newcommand\imprel[1]{\textsc{ImplicitRelations}}
\newcommand\strategy[1]{\textsc{StrategyQA}}
\newcommand\stqa[1]{\textsc{StgQA}}
\newcommand\creak[1]{\textsc{Creak}}
\newcommand\csqa[1]{\textsc{CSQA2}}
\newcommand\commonsenseqa[1]{\textsc{CommonsenseQA 2.0}}
\newcommand\gpt[1]{\textsc{GPT-3}}
\newcommand\pair[2]{$\langle$\textit{#1}, \texttt{#2}$\rangle$}

\newcommand\commentout[1]{}
\newcommand\mg[1]{\textcolor{blue}{[MG: #1]}}

\newcommand\uk[1]{\textcolor{purple}{[UK: #1]}}

\usepackage{inconsolata}

\title{Inferring Implicit Relations in Complex Questions with Language Models} 

\author{Uri Katz$^{1}$ ~~~~~Mor Geva$^{2}$ ~~~~~
Jonathan Berant$^{1}$ \\
$^1$The Blavatnik School of Computer Science, Tel-Aviv University \\
$^2$Allen Institute for Artificial Intelligence \\
\small{\texttt{\{uri.katz,joberant\}@cs.tau.ac.il}}, \small{\texttt{morp@allenai.org}}
}

\begin{document}
\maketitle
\begin{abstract}
A prominent challenge for modern language understanding systems is the ability to answer implicit reasoning questions, where the required reasoning steps for answering the question are not mentioned in the text explicitly. 
In this work, we investigate why current models struggle with implicit reasoning question answering (QA) tasks, by decoupling inference of reasoning steps from their execution. 
We define a new task of \textit{implicit relation inference} and construct a benchmark, \imprel{}, where given a question, a model should output a list of concept-relation pairs, where the relations describe the implicit reasoning steps required for answering the question.
Using \imprel{}, we evaluate models from the GPT-3 family and find that, while these models struggle on the implicit reasoning QA task, they often succeed at inferring implicit relations. 
This suggests that the challenge in implicit reasoning questions does not stem from the need to plan a reasoning strategy alone, but to do it while also retrieving and reasoning over relevant information.
\end{abstract}

\section{Introduction}
\label{sec:introduction}

A longstanding goal of language understanding  has been to develop systems that can reason, i.e., integrate multiple pieces of information to reach a conclusion \cite{McCarthy1960ProgramsW,clark2021transformers}. This has sparked interest in question answering (QA) benchmarks that require such reasoning \cite{welbl-etal-2018-constructing,yang2018hotpotqa,talmor-berant-2018-web}. One particularly challenging case is 
questions that require \emph{implicit reasoning}, 
that is, where the evidence for answering the question is not mentioned explicitly. 
Consider the question \textit{``Does Santa Claus work during summer?''}. This question requires implicit reasoning since it involves knowing when the holiday associated with Santa occurs, but this is not evident from the question.

\begin{figure}[t]
    \centering
    \includegraphics[scale=0.5]{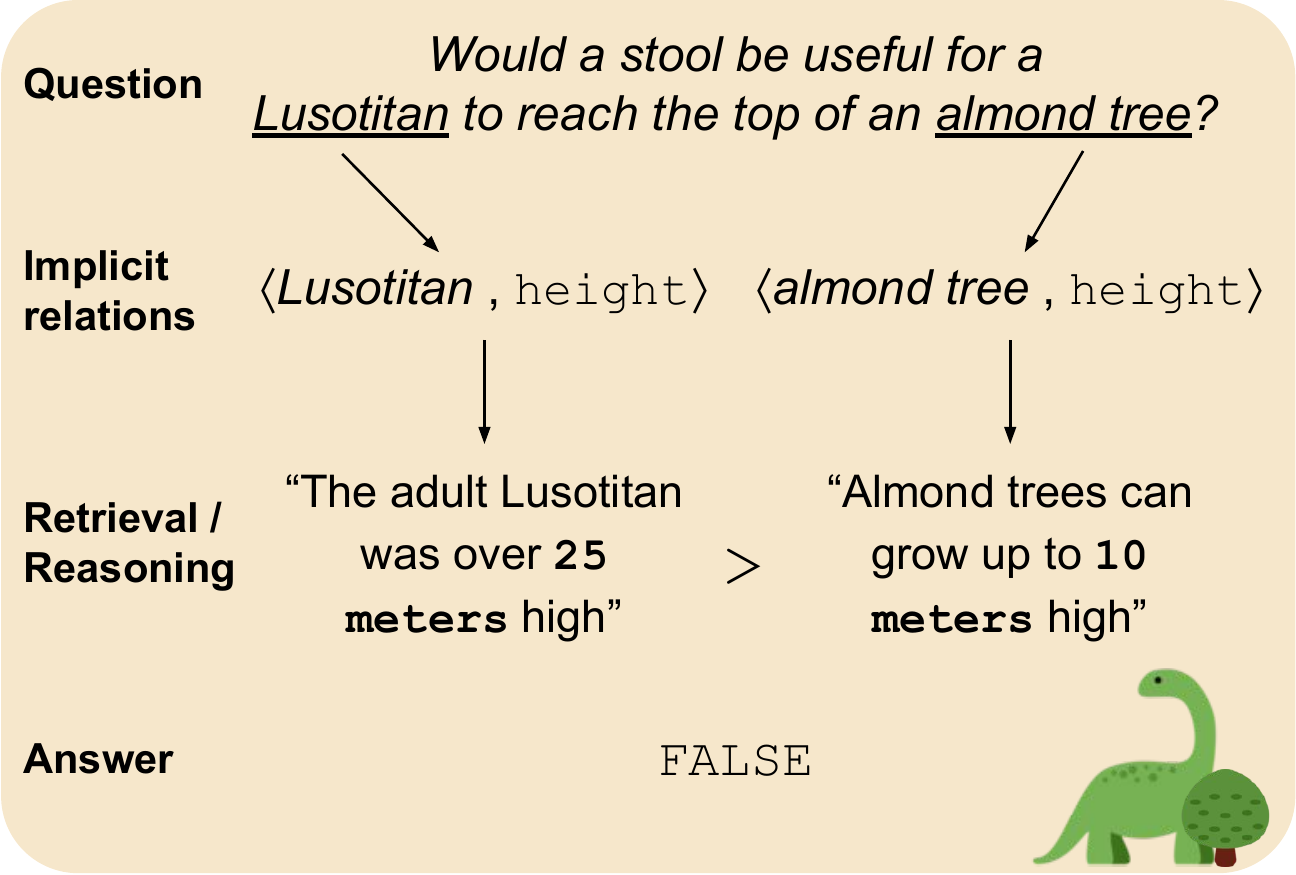}
    \caption{An example \emph{implicit reasoning question}, where answering requires inferring \emph{implicit relations} that are not explicitly mentioned. Besides implicit relations, answering the question also requires reasoning over the relevant retrieved facts. In this work, we focus on the first step of inferring implicit relations.
    } 
    \label{Figures:Figure 1}
\end{figure}

\begin{table*}[t]
    \centering
    \footnotesize
    \begin{tabular}{l p{11.6cm} l}
         \textbf{Source Dataset}  & \textbf{Question and Implicit Relation Pairs} & \textbf{Answer} \\ \toprule
         \multirow{4}{*}{\strategy{}}  & \textit{Did the 40th president of the United States forward lolcats to his friends?} 
         \newline \pair{40th president of the United States}{year of death} , \pair{lolcats}{year of creation} &  \multirow{2}{*}{\texttt{False}}\\\cmidrule{2-3}

         &\textit{Could \$1 for each 2009 eclipse buy a copy of TIME magazine in 2020?}
         \newline \pair{2009 eclipse}{number} , \pair{TIME magazine in 2020}{retail price}&\multirow{2}{*}{\texttt{True}}\\ \midrule

         \multirow{4}{*}{\creak{}} & \textit{Aziz Ansari has never performed in front of a crowd.}
         \newline \pair{Aziz Ansari}{profession}& \multirow{2}{*}{\texttt{False}}\\\cmidrule{2-3}
         
          &\textit{Pantera made music with distorted guitars.} \newline \pair{Pantera}{music genre} & \multirow{2}{*}{\texttt{True}}\\ \midrule
          
          \multirow{4}{*}{\textsc{CSQA}2.0} & \textit{None of the mail in a person's Yahoo inbox has a stamp on it.}
          \newline \pair{Yahoo inbox}{type of mailbox} & \multirow{2}{*}{\texttt{True}}\\\cmidrule{2-3}
          & \textit{If you play a cello you cannot join the marching band.}
          \newline \pair{cello}{playing posture}&  \multirow{2}{*}{\texttt{True}}\\ \bottomrule
    \end{tabular}
    \caption{Example annotations of concept-relation pairs from \imprel{} along with the question source dataset and answer. Each source exhibits different facets of implicit reasoning questions.}
    \label{table:datasets}
\end{table*}

Recent advances in QA \cite{Tafjord2021Macaw,Lourie2021UNICORNOR} have steered attention towards implicit reasoning QA benchmarks such as \strategy{} \cite{geva2021did}, \textsc{OpenCSR} \cite{lin-etal-2021-differentiable}, \commonsenseqa{} \cite{talmor2021commonsenseqa}, \creak{} \cite{onoe2021creak}, and \textsc{RealFP} \cite{kalyan-etal-2021-much}, which span a wide range of domains and reasoning skills. 
Still, implicit reasoning remains an open challenge, even for large language models (LMs) such as GPT-3 and PaLM \cite{bigbench,talmor2021commonsenseqa,rae2021scaling,chowdhery2022palm}.

Answering implicit reasoning questions can be viewed as a two-step process:
(a) inferring simple sub-questions necessary for answering the question, and (b) retrieving the relevant knowledge pieces (i.e., answering sub-questions) and reasoning over them to derive the answer.
Figure~\ref{Figures:Figure 1} illustrates this decoupling. To answer the shown question, we need to use knowledge about the \emph{Lusotitan dinosaur} and \emph{almond trees} to infer that the relevant sub-questions concern their \emph{heights}. We refer to the relation \emph{height}, which is not mentioned in the question as an \emph{implicit relation}. Once implicit relations are inferred, we can retrieve the relevant facts and deduce that the answer is `False', as a Lusotitan is much higher than an almond tree.

In this work, we put a spotlight on \emph{implicit relations} and investigate the ability of language models to infer them as a necessary (albeit insufficient) step for answering implicit reasoning questions.
We first define implicit relations, and show that they can be reliably 
annotated through crowdsourcing (example annotated implicit relations are in Figure~\ref{Figures:Figure 1} and Table~\ref{table:datasets}). To show implicit relations are common, we curate and annotate implicit reasoning questions from three existing datasets, \strategy{}, \creak{}, and \commonsenseqa{}, which results in \imprel{}, a new evaluation benchmark containing 615 questions and 2,673 annotations of implicit relations. 

We use our benchmark to evaluate the ability of large LMs to infer implicit relations, since they are known to acquire substantial amounts of knowledge and common sense with scale \cite{roberts-etal-2020-much,Liu2021GeneratedKP,smith2022using}, but struggle with implicit reasoning questions. Specifically, we evaluate models from the GPT-3 family using \emph{in-context learning}, where the model is fixed and only a few examples are given as context.

We find that large LMs perform well on this task, with a 175B parameter model recovering 0.53-0.59 of the implicit relations across datasets, outperforming a baseline by 21-40 points.
This is robust across methods for sampling in-context examples, and even in cross-data scenarios where in-context examples are sampled from a \emph{different} dataset than the target question. However, inferring implicit relations does \emph{not} improve accuracy on the downstream QA task, even when gold relations are provided.
This suggests that the challenge of implicit reasoning questions is not primarily due to implicit relation inference, but possibly due to the need to also retrieve information and reason over it.

To conclude, in this work we propose the notion of \emph{implicit relations}, and construct the \imprel{} evaluation benchmark for testing the ability of models to infer them from questions. We evaluate large LMs and show that they infer implicit relations fairly well, while still falling short of answering implicit reasoning questions. Our work facilitates future work on improving implicit relation inference, and sheds light on the factors relevant for developing models that can handle implicit reasoning.
From a broader perspective, our work joins recent community efforts to highlight the ubiquity of missing and implicit elements in natural language \cite{,cheng-erk-2018-implicit,pyatkin-etal-2020-qadiscourse,tne}.
\footnote{Our benchmark \imprel{} and relevant code can be downloaded from \url{github.com/katzurik/ImplicitRelations}.}

% =========

\section{Implicit Relations}
\label{sec:implicit_relations}

\begin{figure}[t]
    \centering
    \includegraphics[scale=0.55]{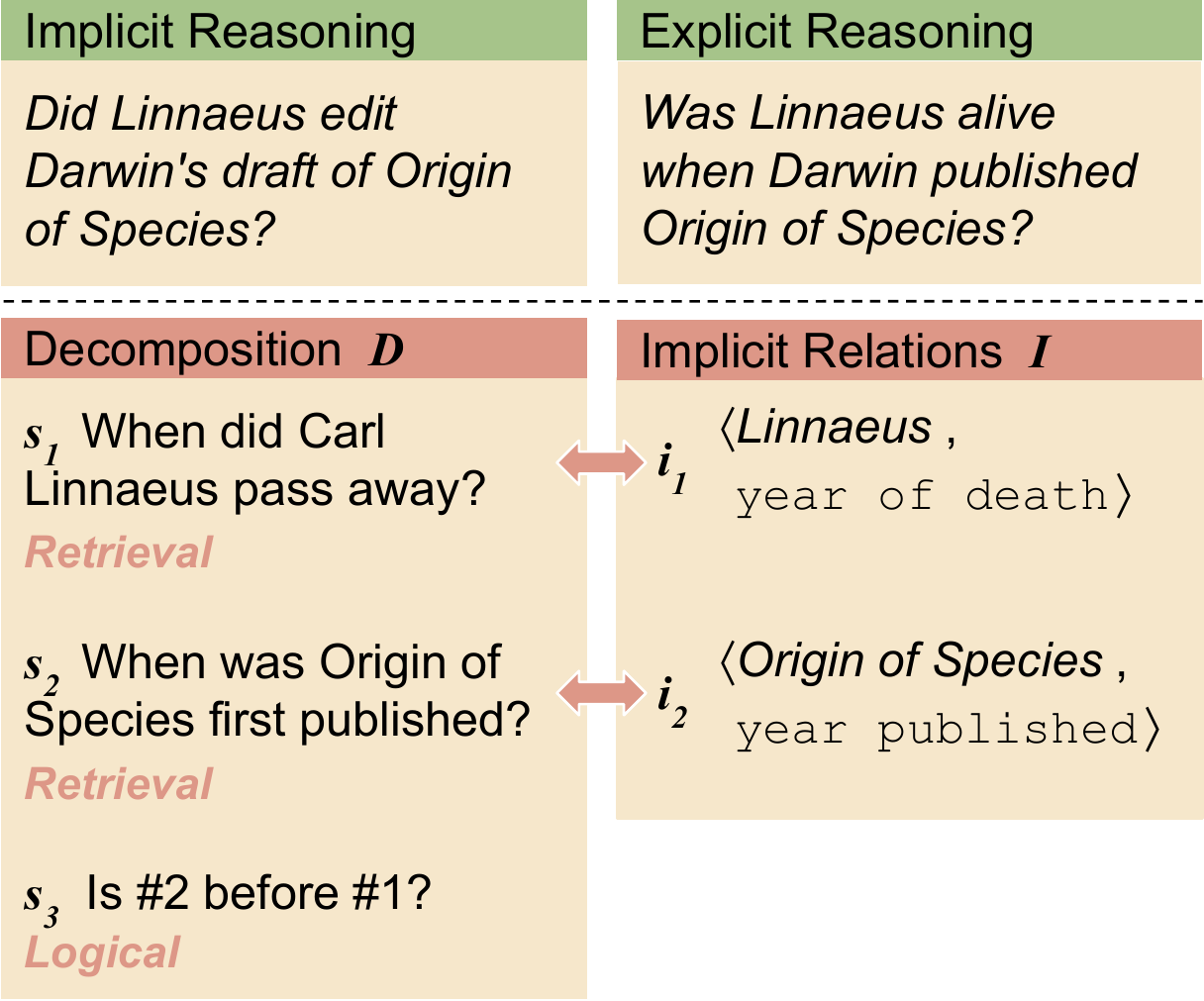}
    \caption{An example of explicit and implicit reasoning questions that share the same question decomposition, along with the implicit relations derived from the retrieval steps in the decomposition.
    } 
    \label{Figures:implicit_relation_figure2}
\end{figure}

We now define the notion of implicit relations in the context of complex question answering.
%\mg{I think we should have a figure with an explicit complex question, its implicit version, the (same) decomposition to reasoning steps, the corresponding implicit relations, and answer.} \jb{yeah I think I already proposed that as something - we need a running example for all the various parts we define here that will make everything just very easy to understand}

%\paragraph{Implicit reasoning QA}
Complex questions are questions that require multiple steps of reasoning in order to be answered \cite{yang2018hotpotqa, talmor-berant-2018-web,welbl-etal-2018-constructing,khot-etal-2021-text}.
For example, the question \emph{``Was Linnaeus alive when Darwin published Origin of Species?''} involves fetching two dates and then comparing them (Figure~\ref{Figures:implicit_relation_figure2}).
%.The question explicitly specifies the steps that need to be followed in order to answer it. 
A prominent challenge in complex QA, that attracted substantial attention recently \cite{mihaylov-etal-2018-suit,khashabi-etal-2020-unifiedqa,lin-etal-2021-differentiable,geva2021did,yasunaga-etal-2021-qa,Wei2022ChainOT}, is cases
where the reasoning steps are \emph{implicit} and should be inferred from the question. For instance, \textit{``Did Linnaeus edit Darwin's draft of Origin of Species?''} involves the same reasoning steps, but they are not mentioned explicitly. Thus, the former question is an \emph{explicit} reasoning question, while the latter is an \emph{implicit} one (Figure~\ref{Figures:implicit_relation_figure2}).

%\paragraph{Question decomposition}
\newcite{wolfson2020break} proposed QDMR as a meaning representation for complex questions, where a complex question $q$ is decomposed into a sequence of $m$ reasoning steps $D = (s_1, \dots, s_m)$, and each step $s_i$ corresponds to a simple natural language question. Answering the simple questions one-by-one yields the final answer (see decomposition in Figure~\ref{Figures:implicit_relation_figure2}). \newcite{geva2021did} collected decompositions for implicit reasoning questions as part of the \strategy{} dataset, where importantly, inferring the sequence of reasoning steps is challenging due to their implicit nature. In addition to generating effective decompositions for implicit reasoning questions, we find an additional challenge in evaluating these decompositions when represented as a sequence of sub-questions.
%Previous work \cite{wolfson2020break,perez-etal-2020-unsupervised}  \jb{text modular qa?} proposed to represent a complex question as a sequence of reasoning steps, each corresponding to a simple question. Executing this decomposition to reasoning steps over a given context yields intermediate answers that can be composed together to derive the final answer. 
%A decomposition of the previous example would be \mg{write decomposition}.
%Question decompositions have also been collected for implicit reasoning questions, as part of the \strategy{} dataset \cite{geva2021did}. 
Specifically, \newcite{geva2021did} distinguished two types of reasoning steps in a decomposition -- \emph{retrieval steps}, which require retrieval of facts ($s_1$ and $s_2$ in Figure~\ref{Figures:implicit_relation_figure2}), and \emph{logical steps},
which perform logical reasoning over previous results ($s_3$ in Figure~\ref{Figures:implicit_relation_figure2}).

%\paragraph{Implicit relations}
In this work, we observe that a key ingredient 
in inferring decompositions is to identify the \emph{implicit relations} that are necessary for answering the question.
% Concretely, in a question decomposition, each retrieval step 
Concretely, each retrieval step in a question decomposition can typically be represented as concept-relation pair $\langle c, r \rangle$, where $c$ is a sequence of tokens from the question that refer to a concept, and $r$ is a relation of that concept. For example, 
the concept $c$ in step $s_2$ in Figure~\ref{Figures:implicit_relation_figure2} is \emph{``Origin of Species''}, and the relation $r$ is its publication year.

%we propose that answering implicit reasoning questions can be viewed as a two-step process of (a) inferring the question decomposition to reasoning steps, and (b) executing the reasoning steps to derive the answer. 
%Moreover, we observe that a key ingredient of inferring the question decomposition involves identifying implicit relations that are necessary for answering the question.
%Concretely, retrieval decomposition steps often can be represented as concept-relation pairs $\langle c, r \rangle$, where $c$ is a concept from the question and $r$ is a relation or property of that concept. For example, \mg{example that demonstrates the correspondence between decomposition steps and implicit relations}.

Based on this observation, we provide the following definition for implicit relations in complex QA. 
Let $q$ be an implicit reasoning question, and denote by $D =(s_1, ..., s_m )$ its decomposition into a sequence of $m$ reasoning steps. Let $\{s_{i_1}, \dots, s_{i_n}\}$ be the subset of retrieval steps in the decomposition $D$.
We define the implicit relations for answering $q$ as the set $\mathcal{I} = \{\langle c_1, r_1 \rangle, ..., \langle c_n, r_n \rangle \}$ of concept-relation pairs, where each concept-relation pair corresponds to a particular retrieval step.
%derived from retrieval steps in $D$. 
%Table~\ref{table:datasets} show examples of questions and their implicit relations.
%\mg{maybe refer to Table~\ref{table:datasets}}

In the next sections, we will use this definition to construct a task for probing the ability of models to infer implicit relations (\S\ref{sec:dataset}-\S\ref{sec:infer_implicit_relation}), and investigate why they struggle on the downstream QA task (\S\ref{sec:qa_performance}).

\section{The \imprel{} Benchmark}

In this section, we describe the process of creating \imprel{}, a benchmark for evaluating the ability of models to infer implicit relations.

\label{sec:dataset}
\subsection{Data Collection}
We curate questions that require inferring implicit relations from three recent datasets:
\begin{itemize}
[leftmargin=*,topsep=3pt,itemsep=3pt,parsep=0pt]
    \item \textbf{\strategy{}} \cite{geva2021did}: A dataset of yes/no questions that require implicit multi-step reasoning. \strategy{} (\stqa{}) questions can be answered from Wikipedia, and are diverse in terms of the required reasoning skills and question topic. Large language models, such as GPT-3 \cite{brown2020language}, were shown to struggle on \stqa{} \cite{bigbench}.
    
    \item \textbf{\creak{}} \cite{onoe2021creak}: A dataset containing true/false statements that require common sense and knowledge about real-world entities.
    %Claims are curated by sampling a large number of entities with varying degrees of popularity, 80\% of which are named entities.  \cite{onoe2021creak} \jb{what are entities that are not named entities? Where is it sampled from?}
    
    \item \textbf{\csqa} \cite{talmor2021commonsenseqa}: A dataset containing yes/no questions and true/false statements. \csqa{} questions involve generic commonsense reasoning. Most questions do not require knowledge about particular entities.
\end{itemize}
Examples from each dataset are shown in Table~\ref{table:datasets}.
% To find questions that require inferring implicit relations, we curate them from three source datasets: \strategy{} \cite{geva2021did}, \creak{} \cite{onoe2021creak}, and \csqa{} \cite{talmor2021commonsenseqa}. We briefly describe these datasets (see examples in Table~\ref{table:datasets}).
% \begin{itemize}
% [leftmargin=*,topsep=3pt,itemsep=3pt,parsep=0pt]
%     \item \textbf{\strategy{}}: A dataset of yes/no questions that require implicit multi-step reasoning. \strategy{} questions can be answered from Wikipedia, and are diverse in terms of the required reasoning skills and question topic. Large language models, such as GPT-3 \cite{brown2020language}, were shown to struggle on \strategy{} \cite{bigbench}.
    
%     \item \textbf{\creak{}}: A dataset containing true/false statements that require common sense and knowledge about real-world entities \cite{onoe2021creak}.
%     %Claims are curated by sampling a large number of entities with varying degrees of popularity, 80\% of which are named entities.  \cite{onoe2021creak} \jb{what are entities that are not named entities? Where is it sampled from?}
    
%     \item \textbf{\csqa}: A dataset containing yes/no questions and true/false statements. \csqa{} questions involve generic commonsense reasoning. Most questions do not require knowledge about particular entities \cite{talmor2021commonsenseqa}.
% \end{itemize}

Collecting questions from three sources serves two purposes. First, it demonstrates that inferring implicit relations is necessary for many question types: in multi-step questions (\stqa{}) and single-step questions (\creak{}), in entity-focused questions (\creak{}) and generic common sense questions (\csqa{}). Second, these datasets were created using different protocols: \stqa{}{} and \csqa{} use a model-in-the-loop during data collection, while \creak{} does not. \stqa{}{} and \creak{} ask annotators to author questions freely, while \csqa{} employs a gamification mechanism. Having questions from different data collection pipelines increases the likelihood that empirical conclusions are not tied to a particular dataset.

%\jb{I felt this paragraph was really unclear, so I wrote another above, let me know if you think it's probelmatic}
%The three data sources contribute questions that use different reasoning skills, ranging from simple and straightforward commonsense knowledge questions to complex, multi-step strategy questions. As the current SOTA few-shot setting for each data source is well below human performance, it is imperative to investigate further into the reasoning mechanism of LMs. Despite the sources diversity, they all contain questions that require one to decipher the implicit reasoning process to get to the answer. We were motivated by the fact that this property of reasoning process is not limited to a single dataset or task and can be observed in QA tasks frequently.

\paragraph{Question curation}
We chose questions that satisfy two properties: (a) answering the question requires inferring an implicit relation, and (b) the question is \emph{feasible}, that is, it can be answered using real-world facts (provided as part of the benchmark in \stqa{} and \creak{}) or using generic common sense (in \csqa{}).
We sampled examples from the training set of each of the three datasets and kept questions that satisfy the two properties. 
% \mg{maybe mention already here how many questions we had at this point.}\uk{Does the wording of the paragraph lead the reader to ask about the ratio of questions with implicit relations out of the total questions? it is something we would like to avoid IMO} 
%jb: below seems unnecessary
%For \strategy{}, we found that almost all questions require inferring implicit relations, and thus we judged feasibility only. 
%\jb{do we want to write something about how many questions in each dataset satisfied the two properties saying that we erred on the side of filtering too much than too little?}

\paragraph{Annotating implicit relations}
%We manually curated samples from \creak{} and \csqa{} by selecting questions that do not explicitly state the reasoning process or do not include any reasoning steps at all. We did not limit our selection to any topic, reasoning skills, or other property other than the implicit nature of the question. We confirmed that all questions are feasible regardless of their complexity. \strategy{} questions were randomly selected and only assessed for feasibility. All the questions were selected from the training  sets of each source. 
We use Amazon Mechanical Turk to annotate implicit relations over curated questions. We qualify 15 crowdworkers to identify concepts, which are token sequences from the question, relevant for answering the question, and the corresponding relation for each concept (see annotation guidelines in Appendix~\ref{sec:mturk}).
% Annotators were given the option to specify up to four different concept-relation pairs per question, but in practice 98.9\% of the annotations consist of $\leq 2$ concept-relation pairs.
Annotators can specify up to four concept-relation pairs per question, but in practice, 98.9\% consist of $\leq 2$ pairs.
Concepts must be extracted directly from the input questions, and relations are phrased using concise natural language phrases. 

For \stqa{} and \creak{}, which often require uncommon knowledge about entities, we provided additional context from the original data source.
% For \stqa{}, we provided facts the original question author wrote along with the full question decomposition. For \creak{}, we provided an explanation for why the claim is true or false, written by the original claim author. 
For \stqa{}, we provided facts along with the full question decomposition. For \creak{}, we provided an explanation for why the claim is true or false.

We collected 5 annotations per example in \creak{} and \csqa{}, and 3 annotations in \stqa{}. Due to the availability of facts and question decompositions, \stqa{} showed high agreement between annotators (see Table ~\ref{table:datasets_summary}). \creak{} and \csqa{} showed more variability, and thus, we collected more annotations per example.
To ensure quality, we manually verified all examples created during the annotation process, filtering out annotations that do not fit the task requirements. 

\subsection{Data Analysis}

% \begin{table*}[t]
%     \centering
%     \footnotesize
%     % \begin{tabular}{p{2.4cm}|p{1.3cm}|p{1.3cm}|p{1.3cm}|p{1.3cm}|p{1.3cm}|p{1.3cm}}
%     \begin{tabular}{llllrrr}
%          \textbf{Source dataset}  & \textbf{Questions} & \textbf{Unique} & \textbf{Entity-relation} & \textbf{Entity}  & \textbf{Implicit relation} & \textbf{Annotators} \\
%          &  & \textbf{entities} & \textbf{pairs} & \textbf{agreement}  & \textbf{similarity} &  \\ \toprule
%          \strategy{}  & 201 & 399 & 1139 & \%84 & 0.59 & 3 \\
%          \creak{} & 205 & 309 & 1272 &\%74 & 0.52 & 5 \\ 
%          \csqa{}  & 209 & 303 & 1285 & \%75 & 0.56 & 5 \\ \bottomrule
%     \end{tabular}
%     \caption{--r-r-r--r-- 
%     \jb{need to say what numbers are some over all annotators}}
%     \label{table:datasets_summary}
% \end{table*}

\begin{table}[t]
    \setlength\tabcolsep{3.0pt}
    \centering
    \footnotesize
    \begin{tabular}{lccc}
        & \stqa{} & \creak{} & \csqa{} \\ 
        \midrule
        \# of questions & 201 & 205 & 209 \\
        \# of unique concepts & 399 & 309 & 303 \\
        \# of concept-relation pairs & 1139 & 1272 & 1285 \\
        \midrule
        Concept agreement & 87\% & 71\% & 74\% \\
        \midrule
        Relation agreement: \\
        Lexical variability & 85\% & 65\% & 80\% \\
        Multiple reasoning paths & 15\% & 35\% & 20\% \\
        \bottomrule
    \end{tabular}
    \caption{Statistics on \imprel{}.}
    \label{table:datasets_summary}
\end{table}

Table~\ref{table:datasets_summary} provides statistics on the collected data.
\imprel{} consists of 615 annotated questions,
$\sim200$ per dataset, where exactly 100 examples from each data source are used as a test set, and the rest are used as a development set. 
Development set examples are in Appendix~\ref{sec:example_appendix}.

% \imprel{} consists of 615 annotated questions,
% % (each with multiple annotations),
% $\sim200$ per dataset, where exactly 100 examples from each data source are used as a test set, and the rest are used as a development set. See Table~\ref{table:datasets_summary} for exact statistics. Additional random examples from the development set are provided in Appendix~\ref{sec:example_appendix}.

%Annotator agreement
\paragraph{Annotator Agreement}
\label{par:annotator_agreement}
We manually analyzed 20 random examples from each data source to evaluate agreement between annotators for concepts and relations. We declared concept agreement when at least three annotators identified the same concept in an example. We found that annotators agreed on 77\% of the concepts, and less than 10\% of concepts were extracted only by a single annotator. See Table~\ref{table:datasets_summary} for break-down by data source.

Assessing agreement on relations is more challenging, since relations are short phrases that can differ lexically. We marked for each example whether annotated relations differed only lexically or due to multiple reasoning strategies for answering the question. In 76\% of the examples, the relations across \emph{all} annotators were identical or differed lexically, e.g., the relations \emph{``average price''} and \emph{``cost''}. In 24\% of the examples, multiple reasoning strategies were used, which would result in different reasoning and retrieval steps (see Table~\ref{table:datasets_summary}).

Overall, our analysis suggests that implicit relations can be annotated reliably.

\section{Experimental Setting}
\label{sec:experimental_setting}

We now turn to evaluate the ability of large LMs to infer implicit relations in questions. To this end, we use examples from \imprel{} in a few-shot \emph{in-context learning} setting \cite{brown2020language}, where given several input-output examples and a test input, the LM is expected to generate the required output.
We focus on this setup following the recent progress in in-context learning, specifically for tasks that involve general commonsense reasoning \cite{da2021analyzing,chowdhery2022palm}.

\subsection{Task and Model Specification}

% We study the ability of large LMs to infer implicit relations in the few-shot setting through \emph{in-context learning} \cite{brown2020language}, where given a few annotated examples and a test input, the LM is asked to generate the required output. We focus on this setup due to the paucity of annotated data and recent progress in in-context learning, specifically for tasks that involve general commonsense reasoning \cite{chowdhery2022palm,da2021analyzing}.

Given a test question $q^*$, we prepend $k$ annotated examples $\{\langle q^{(i)}, \mathcal{I}^{(i)} \rangle\}_{i=1}^k$ to it, where $q^{(i)}$ is the $i$-th question in this set and $\mathcal{I}^{(i)} = \{ \langle c^{(i)}_1, r^{(i)}_1 \rangle, ..., \langle c^{(i)}_{n_i}, r^{(i)}_{n_i} \rangle \}$ is the set of corresponding concept-relation pairs.
%\mg{maybe we can find a better notation}.
Specifically, we use the following input format:
\setlength{\jot}{0pt}
\begin{align*}
&\textbf{Question:} \;\; q_1 \\ 
&\textbf{Implicit Reasoning:} \;\; \langle c^{(1)}_1, r^{(1)}_1 \rangle, ..., \langle c^{(1)}_{n_1}, r^{(1)}_{n_1} \rangle \\
&\dots \\
&\textbf{Question:} \;\; q_k \\
&\textbf{Implicit Reasoning:} \;\; \langle c^{(k)}_1, r^{(k)}_1 \rangle, ..., \langle c^{(k)}_{n_k}, r^{(k)}_{n_k} \rangle \\
&\textbf{Question:} \;\; q^* \\
&\textbf{Implicit Reasoning:}
\end{align*}
Example inputs are given in Appendix~\ref{sec:example_appendix}.
The prefixes ``Question'' and ``Implicit Reasoning'' were chosen arbitrarily and remained fixed throughout all experiments.
For each test question, $k$ examples are randomly sampled from the development set, and for each example, a single random annotation is selected. In our experiments, we use $k=16$.\footnote{We experimented with $k \in \{8, 16, 32$\} but found it does not dramatically change performance, see Appendix~\ref{sec:prompt_size_appendix}.}

\paragraph{Models}
We evaluate models from the GPT-3 family,\footnote{We use the API at \url{https://openai.com/api/}.} which are known to exhibit broad factual knowledge \cite{brown2020language}.
% We perform all experiments with LMs from the GPT-3 family \cite{brown2020language}, using the OpenAI API.\footnote{\url{https://openai.com/api/}.} 
In particular, we use \texttt{text-davinci-002}, a 175B-parameter LM, which to the best of our knowledge was not trained on any of the target benchmarks. Furthermore, to assess the scaling behaviour of models, we experiment with other models from the GPT-3 family, see  \S\ref{subsec:model-scale} for details.
%For the model scales experiment, to make a comparison between different model sizes that is as accurate as possible, we used the \texttt{davinci} model rather than the \texttt{text-davinci} model, which is not yet available for all other sizes \mg{Not sure but maybe we should mention this only when we talk about this experiment, right now it's a bit out of context}. The following models were used  \texttt{ada}, \texttt{babbage}, \texttt{curie}, and \texttt{davinci} which are assumed to correspond to 350M, 1.3B, 6.7B, and 175B parameter models, respectively \cite{gao2021sizes}.
In all experiments, outputs are predicted using greedy decoding.% and are limited to 50 tokens.
% was used to sample the models. \mg{do we limit the number of generated tokens?}\uk{yes, to 50 ,but it is stopping by itself because of the stopping criteria for newline  which 100\% of the cases stopped the generation  }

\paragraph{Baseline} 
To account for correlations originating from the concepts that appear in the question or reasoning shortcuts, we define a \textit{`Concept-only'} baseline, where instead of testing whether the model can infer the implicit relations from the full question, we test its ability to infer them from the set of \emph{gold} concepts that appear in the question. For this baseline, we use the same inputs as before, but replace every question $q_i$ and test question $q^*$ with its set of annotated concepts.
\begin{align*}
\textbf{Question:} \;\; q_i \;\rightarrow\; \textbf{Question:} \;\; c^{(i)}_1 \;;\; ... \;;\; c^{(i)}_{n_i}
\end{align*}
While the identity of the gold concepts provides useful information for inferring implicit relations, we expect models that have access to the full question to perform better.

\subsection{Evaluation}
Inferring implicit relations involves identifying concepts and relations. We now define evaluation metrics for this task. 
% We define the following evaluation metrics.

\paragraph{Concept extraction} Our output is a set of conept-relation pairs. 
Let $\mathcal{C}_\text{pred}$ be the set of concepts predicted by the LM, and let $\mathcal{C}_\text{gold}^i$ be the gold set annotated by annotator $i$.
Given that annotated concepts are tokens from the original question, we can use edit-distance
%\footnote{\url{https://github.com/seatgeek/thefuzz}.} 
to match each predicted concept $c \in \mathcal{C}_\text{pred}$ with a concept from $\mathcal{C}_\text{gold}^i$, declaring a match when the edit distance of the best match is above 0.8. Following concept matching, we compute recall and precision in the typical fashion and take the $\max$ over all annotators.
A post-hoc manual analysis validated that no incorrect concept matching occurred 
% in the three data sources 
due to the use of edit distance.

%We assessed the concept extraction recall and precision for the model-inferred concepts compared to the annotated concepts. Based on the fact that the annotated concepts are a span of the original question, inferred concepts are expected to also be derived directly from the input question. We used edit distance\footnote{\url{https://github.com/seatgeek/thefuzz}.} to match the inferred concepts to the annotated ones. A correct concept extraction was declared for all matches with a score above 0.80. A post-hoc manual evaluation found that no incorrect concept matching occurred for any of the three data sources. As there are multiple annotations per question, we compute concept extraction recall and precision for each annotation. We calculate the recall for each annotation and take the annotation with the highest recall score as the final recall and precision.\uk{is it understood? we take the recall for the maximum score, and use that annotation for precision aswell }

\begin{figure}[t]
    \centering
    \includegraphics[scale=0.9]{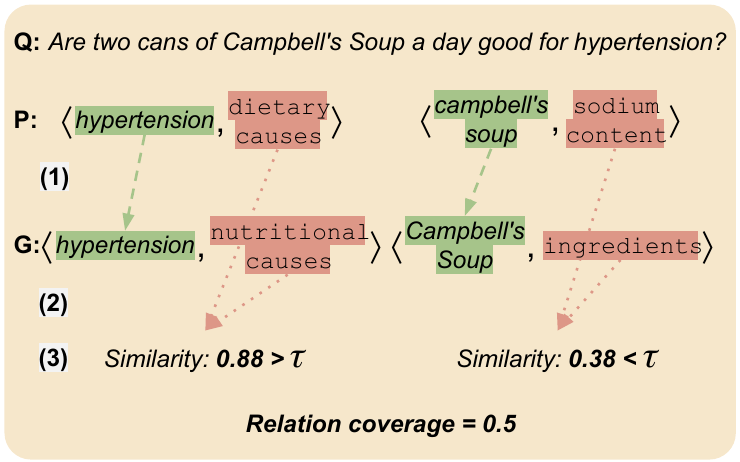}
    \caption{
    Given predicted concept-relation pairs (P) and gold concept-relation pairs (G), we evaluate relations by 
    aligning predicted and gold concepts using edit distance \textbf{(1)}, and computing the cosine similarity between matched relation embeddings \textbf{(2)}. Relation coverage is the fraction of gold relations with similarity $>\tau$ \textbf{(3)}. %Note that this example contains a false negative since \emph{sodium content} is a valid implicit relation.
    } 
    \label{Figures:relation_matching}
\end{figure}

\paragraph{Relation coverage}
Since relations are short phrases with high lexical variability, string matching is not a viable strategy. 
To overcome this, we leverage our ability to align concepts
and use relation embeddings rather than relation strings.
%and use them as pivots for matching relations. Moreover, we compute similarity between relation embeddings rather than relation strings.
Figure~\ref{Figures:relation_matching} depicts our evaluation procedure for two sets of predicted ($\mathcal{R}_\text{pred}$) and annotated ($\mathcal{R}_\text{gold}$) relations.
%\mg{again, we need to define the notation of relations earlier}.
First (Figure~\ref{Figures:relation_matching}, step 1), we align predicted and gold concept-relation pairs, using the concepts (as done for concept evaluation). 
Then (Figure~\ref{Figures:relation_matching}, step 2), we embed every relation phrase, using the sentence transformer \texttt{all-mpnet-base-v2}
%\footnote{\url{https://www.sbert.net/docs/pretrained_models.html}} 
\cite{reimers-2019-sentence-bert}, and compute cosine similarity between the embeddings of matched relations (defined it as zero if no relation was matched).
Last (Figure~\ref{Figures:relation_matching}, step 3), we consider a gold relation $r_\text{gold}$ \emph{covered} if cosine similarity is higher than a threshold $\tau$, and compute 
\emph{relation coverage}, that is, the fraction of gold relations that were covered. 
With this procedure, we evaluate model predictions against each annotation and take the maximum as the final relation coverage.

% Assessing predicted relations is more challenging since relations are short phrases that exhibit high lexical variability, and thus, evaluating with string matching is not a viable strategy. Instead, we perform the following procedure for each annotation (illustrated in Figure~\ref{Figures:relation_matching}):
% First, we match annotated relations to predicted ones by performing concept matching as described above. Then, for each of the gold relations from annotator $i$, $r_\text{gold} \in \mathcal{R}_\text{gold}^i$, we find the matched relation $r_\text{pred}$, if it exists, using the concept matching. We then embed $r_\text{gold}$ and $r_\text{pred}$ using the sentence transformer \texttt{all-mpnet-base-v2}\footnote{\url{https://www.sbert.net/docs/pretrained_models.html}} \cite{reimers-2019-sentence-bert} and compute their cosine similarity (defining it as zero if no relation was matched during concept matching). We then consider a gold relation $r_\text{gold}$ to be successfully matched if the cosine similarity is higher than a threshold $\tau$, and compute 
% \emph{relation recall}, that is, the fraction of gold relations that were successfully matched. The final relation recall takes a $\max$ over all annotators. \uk{the definition here for relation recall is what i called "relation match" , is relation recall sounds better ? }
We focus on coverage (rather than precision) since we care about whether a model can reveal implicit relations, but predicting additional ones is mostly harmless. Moreover, since we use in-context learning, the average number of concept-relation pairs generated is similar to the average number of gold concept-relation pairs (Table \ref{table:mean_concept_relation_pairs}).

% \begin{table}
%     \centering\footnotesize
%     % \begin{tabular}{l|c|c|c|c}
%     \begin{tabular}{lll}
%          \toprule
%          &\textbf{Mean number of gold} & \textbf{Mean number of model generated}  \\
%          &\textbf{concept-relation pairs} & \textbf{concept-relation pairs}\\ \midrule
%          \stqa{}  & 1.9 & 2.0$\pm$0.04   \\ 
%          \creak{} & 1.2 & 1.2$\pm$0.02  \\ 
%          \csqa{}  & 1.2 & 1.2$\pm$0.01   \\ \bottomrule
%     \end{tabular}
%     \caption{The mean number of concept-relation pairs for the annotated development set in each data source. For the model generated pairs, we report an  average over 3 seeds}
%     \label{table:mean_concept_relation_pairs}
% \end{table}

\begin{table}
    \setlength\tabcolsep{3.5pt}
    \centering\footnotesize
    \begin{tabular}{llll}
         & \stqa{} & \creak{} & \csqa{} \\ \midrule
         \# of gold pairs & 1.9 & 1.2 & 1.2 \\
         \# of generated pairs & $2.0\pm0.04$ & $1.2\pm0.02$ & $1.2\pm0.01$  \\
         \bottomrule
    \end{tabular}
    \caption{The mean number of concept-relation pairs in the development set of each data source. For model-generated pairs, we report an average over 3 seeds.}
    \label{table:mean_concept_relation_pairs}
\end{table}

To set a threshold $\tau$ on relation embedding similarity, we annotated whether a predicted relation is semantically equivalent to a matched gold relation for 100 development examples. We choose $\tau=0.51$, which results in a 5\% false-positive rate (predicting that two different relations are equivalent) and 12\% false-negative rate (predicting that two equivalent relations are different). All reported results are an average over three random seeds.

\section{Large LMs Can Infer Implicit Relations}
\label{sec:infer_implicit_relation}

\begin{table}
    \centering\footnotesize
    % \begin{tabular}{l|c|c|c|c}
    \begin{tabular}{llllr}
        %  \toprule
         &  &\textbf{Concept} & \textbf{Concept}  & \textbf{Relation}  \\
         &&\textbf{Recall} & \textbf{Precision} & \textbf{Coverage}  \\ \midrule
         \multirow{2}{*}{\stqa{}} & CO  & 0.99$\pm$0.01 & 0.95$\pm$0.02 & 0.32$\pm$0.02   \\ 
         & FQ & 0.97$\pm$0.01 & 0.89$\pm$0.02 & 0.53$\pm$0.02   \\ \midrule
         \multirow{2}{*}{\creak{}} & CO & 1$\pm$0.0 & 0.96$\pm$0.0 & 0.33$\pm$0.03  \\ 
         & FQ & 0.98$\pm$0.0 & 0.95$\pm$0.01 & 0.54$\pm$0.05  \\ \midrule
         \multirow{2}{*}{\csqa{}} & CO  & 1$\pm$0.0 & 0.98$\pm$0.02 & 0.19$\pm$0.02   \\
         & FQ & 0.93$\pm$0.02 & 0.94$\pm$0.01 & 0.59$\pm$0.01   \\\bottomrule
    \end{tabular}
    \caption{Test set performance for concept-only (CO) and full-question (FQ) for all datasets.}
    \label{table:implicit_relation_eliciting}
\end{table}

Table~\ref{table:implicit_relation_eliciting} shows results on implicit relation inference.
First, the model successfully identifies the relevant concepts in the question, achieving high concept recall and precision across all datasets. This is not limited to named entities but is also achieved when concepts are more abstract, as in \csqa{} (Table \ref{table:datasets}). 
More importantly, \gpt{} infers the implicit relations well, achieving relation coverage scores of 0.53, 0.54, and 0.59 on \stqa{}, \creak{} and \csqa{}, respectively. Moreover, a model exposed to the full question dramatically outperforms a model exposed only to the gold concepts by 21, 21, and 40 points on the three datasets. This indicates that concepts contain relevant information, but access to the full question allows the LM to infer the reasoning strategy and in turn the implicit relations. We provide examples for predicted vs. gold concept-relation pairs along with additional qualitative analysis in  Appendix~\ref{sec:example_appendix}.

Next, we perform additional experiments to (a) further substantiate the ability of LMs to infer implicit relations (\S\ref{subsec:in-context-examples}), and (b) test the effect of model scale on performance (\S\ref{subsec:model-scale}).

\subsection{Effect of In-Context Examples}
\label{subsec:in-context-examples}

While the aforementioned results are encouraging, there are two potential (non-disjoint) causes for them: (a) the LM ``understands'' the task of implicit relation inference, or (b) the LM observes in-context examples and uses them to guess implicit relations for the target question (``soft copying'').
We study the effect of these causes.

%In spite of the fact that \imprel{} includes questions with a wide variety of reasoning types, individual questions may share a common reasoning process or a strategy. As a consequence, the model performance on our task may be overestimated, which is why we would like to conduct a control experiment.

\paragraph{Similar vs. dissimilar in-context examples}
% As explained, we randomly sample $k$ in-context examples from the development set.
% Recall that each input to the model includes $k$ in-context examples from the development set.
To quantify the effect of in-context examples, rather than choosing them randomly, we use examples that are similar or dissimilar to the target question in terms of their implicit relations. 

We first represent each example as an embedding vector, by (a) concatenating all annotated relations, i.e., $r_1, \dots, r_n$, and computing a vector representation using a sentence transformer, and (b) averaging the embedding vectors of all annotators. 
Then, for each example, we select two sets of in-context examples: (a) \emph{Similar}: the top-$k$ most similar examples (using cosine similarity), and (b) \emph{Dissimilar}: we discard the 33\% most similar examples, and randomly sample from the rest. In both cases, we use gold implicit relations at test time, and thus this experiment is for analysis only.

\begin{table}
    \centering\footnotesize
    % \begin{tabular}{l|c|c|c|c}
    \begin{tabular}{llll}
        %  \toprule
         &  & \textbf{Relation} & \textbf{Copying}   \\
         && \textbf{Coverage} &  \\ \midrule
         \multirow{3}{*}{\stqa{}} & Dissimilar  &  0.46$\pm$0.01 & 0.01$\pm$0.01  \\ 
         & Random &  0.52$\pm$0.02 & 0.13$\pm$0.02 \\
         & Similar &  0.54$\pm$0.02 & 0.31$\pm$0.02  \\
         & Concept-only &  0.28$\pm$0.04  & 0.10$\pm$0.04 \\\midrule
         \multirow{3}{*}{\creak{}} & Dissimilar & 0.60$\pm$0.02 & 0.02$\pm$0.02  \\
         & Random &  0.59$\pm$0.03 & 0.08$\pm$0.02 \\
         & Similar &  0.67$\pm$0.02 & 0.33$\pm$0.03  \\
         & Concept-only &  0.34$\pm$0.02 & 0.13$\pm$0.02 \\\midrule
         \multirow{3}{*}{\csqa{}} & Dissimilar  &  0.63$\pm$0.02 & 0.02$\pm$0.02  \\
         & Random &  0.67$\pm$0.01 & 0.07$\pm$0.02 \\
         & Similar &  0.73$\pm$0.01 & 0.21$\pm$0.03 \\
         & Concept-only &  0.20$\pm$0.02 & 0.10$\pm$0.02
         \\\bottomrule
    \end{tabular}
    \caption{Development set performance. Controlling the set of in-context examples with Dissimilar, Similar, Random relations, and Concept-only baseline.}
    \label{table:prompt_ranker}
\end{table}

%In order to create a controllable experiment that will allow us to control the implicit relations examples in the prompt, for each question we sorted all other questions based on their relation similarity and removed the 33\% most similar questions. Then, to construct the prompt we sampled random questions from this limited pool of questions (66\% of the dataset). 
%As part of our analysis of the effect of different examples in the prompt on the performance of the model, we designed a prompt relation ranker. By using Sentence Transformer~\cite{reimers-2019-sentence-bert}, For each question we calculate the embedding vector of the annotated relations, and averaged it for all annotators, yielding the mean relation vector. Then, in order to rank question by similarity / dissimilarity, we scored the pairwise cosine-similarity between all examples. By using this procedure we created an oracle that can control the prompt based on the annotated implicit relation.

Table~\ref{table:prompt_ranker} shows relation coverage for the different sets of in-context examples and the fraction of cases where one of the implicit relations predicted by the LM is copied from the in-context examples. When \emph{Dissimilar} examples are presented, there is a slight performance degradation, most notably in \stqa{}. However, results are still dramatically higher compared to Concept-only. Moreover, the model succeeds in predicting implicit relations while hardly copying from in-context examples. 

In the \emph{Similar} setting, performance increases across all datasets, along with a much higher rate of copying. This hints that designing methods for retrieving similar prompts can lead to gains in performance \cite{rubin2022epr}.

To further investigate the relation between copying and performance, we label every example for whether the model copies from in-context examples and the coverage of the inferred implicit relations. We then compute the point-biserial correlation \cite{tate1954correlation} to check if copying is correlated with performance and find that correlation is low ($<0.1$ for all datasets),
% (0.05, 0.04, and 0.08 for \strategy{}, \csqa{}, and \creak{}, respectively). 
showing that copying does not explain model performance.

Overall, this experiment suggests that while models can leverage examples from context to improve performance, the LM does more than copy and execute implicit relation inference.

%Results are shown in Figure~\ref{Figures:prompt_ranker}. The evaluation showed that the LM was able to infer implicit relations despite a slight reduction in relation matching. These results suggest that the implicit reasoning capacities of the model are not dependent on a specific context in the prompt. In contrast, we designed prompts that are composed of the 16 most similar implicit relations for every test question, and we found that the relation match score increased for all sources of data. Based on these results, it appears that when the context shares similarities in the type of reasoning or strategy, the model can utilize it to perform the task better. Both similar and dissimilar prompt experiments shed light on the potential benefits of prompt constriction for reasoning and common sense NLP tasks.

% \begin{figure}
%     \centering
%     \includegraphics[scale=0.4]{Figures/prompt_ranker_relation_match.pdf}
%     \caption{\uk{caption}} 
%     \label{Figures:prompt_ranker}
% \end{figure}

\paragraph{Cross-dataset in-context examples}
If LMs can infer implicit relations, we should expect high performance even when in-context examples and target questions are taken from different datasets.

To test this, we evaluate performance on questions from \creak{} and \csqa{} when in-context examples originate from all 3 datasets. Testing on \stqa{} does not work well because the number of implicit relations in an example is typically two, while in \creak{} and \csqa{} it is typically one (see Table~\ref{table:mean_concept_relation_pairs}), and thus the LM output a single implicit relation, leading to poor relation coverage.

Table~\ref{table:cross_dataset} shows that, overall, relation coverage remains high for all sources, suggesting that the LM indeed infers implicit relations regardless of the question and reasoning types in the source dataset.
Concept recall and precision are also relatively stable, except when using \stqa{} for in-context examples, since the model tends to output two concept-relation pairs, reducing  precision. Thus, the LM is sensitive to \emph{the number} of output concept-relation pairs that appear in in-context examples, but succeeds in inferring implicit relations.

\begin{table}[t]
    \setlength\tabcolsep{4.0pt}
    \centering\footnotesize
    % \begin{tabular}{l|c|c|c|c}
    \begin{tabular}{lllr}
         \toprule
         \textbf{Inference/Context}& \textbf{Concept} & \textbf{Concept}  & \textbf{Relation}  \\
         \textbf{Source} &\textbf{Recall} & \textbf{Precision} & \textbf{Coverage}  \\ \midrule

         \creak{}/\stqa{}  & 0.97$\pm$0.0 & 0.69$\pm$0.01 & 0.62$\pm$0.03   \\ 
         \creak{}/\creak{}& 0.93$\pm$0.02 & 0.91$\pm$0.01 & 0.59$\pm$0.03   \\
         \creak{}/\csqa{}& 0.93$\pm$0.01 & 0.90$\pm$0.02 & 0.60$\pm$0.0\\ \midrule
         
         \csqa{}/\stqa{}  & 0.98$\pm$0.01 & 0.77$\pm$0.02 & 0.62$\pm$0.0   \\ 
         \csqa{}/\creak{}& 0.96$\pm$0.02 & 0.94$\pm$0.02 & 0.59$\pm$0.02   \\
         \csqa{}/\csqa{}& 0.95$\pm$0.01 & 0.96$\pm$0.0 & 0.67$\pm$0.01\\\bottomrule
    \end{tabular}
    \caption{Development set performance in the cross-dataset setup where inference on example from one dataset is done with in-context examples from another dataset.}
    \label{table:cross_dataset}
\end{table}

%\paragraph{Implicit relation \mg{prediction/inference?}.}
%  We found that the relation match substantially decreased from 0.52$\rightarrow$0.28. The results show that the concepts themselves contain some information that can be used by the model. However, using concept-information alone is not sufficient for our task, and the model is indeed eliciting the relation from the implicit knowledge of the question. 
%\label{par:model_scales}

\subsection{Effect of Model Size}
\label{subsec:model-scale}

Recent work \cite{kaplan2020scaling,smith2022using,chowdhery2022palm} has shown that reasoning  abilities of LMs improve with model size.
%Large number of studies have indicated that the %reasoning capabilities of pretrained large LMs is increasing as the model size increases \cite{brown2020language,kaplan2020scaling,smith2022using}.
%recently has \cite{chowdhery2022palm} shown that \strategy{} QA performance increases as model scale increases. It is therefore necessary to evaluate the dynamics of our task on the basis of different GPT-3 model scales. 
We evaluate this effect on models from the \gpt{} family: \texttt{ada}, \texttt{babbage}, \texttt{curie}, and \texttt{davinci}, which are estimated to have 350M, 1.3B, 6.7B, and 175B parameters, respectively \cite{gao2021sizes,gpt-neox-20b}.  \texttt{text-davinci}, the model evaluated thus far, is a more recent LM that \cite{ouyang2022training} was trained differently.\footnote{\texttt{text-davinci} has 175B parameters like \texttt{davinci}, but its relation coverage on \stqa{} is higher: 0.43$\rightarrow$0.52. This indicates that its training procedure improves inference of implicit relations.} 

Table~\ref{table:implicit_relation_model_scale} presents results on \stqa{}.
%Increasing model size improves concept extraction recall, which is in line with evidence that broad factual knowledge increases as model size increases \cite{brown2020language,kirstain2021few}. 
Increasing model size improves relation coverage and concept recall, but does not significantly change concept precision.
%Moreover the relation match score increases, again indicating a relationship between model size and reasoning or commonsense capabilities.
Moving from \texttt{curie} to \texttt{davinci} leads to a modest gain in relation coverage. Comparing this to the order of magnitude difference in parameters between \texttt{curie} and \texttt{davinci} suggests that inferring implicit relations does not explain performance improvement in many reasoning and commonsense QA benchmarks. The smallest model, \texttt{babbage}, tends to produce structural errors, indicating it did not properly learn the task.

\commentout{
\begin{figure}[h]
    \centering\tiny
    \includegraphics[scale=0.5]{Figures/model_scales.pdf}
    \caption{\mg{this plot is very wasteful in terms of space. Can we use a table/text only instead? If not, I would just decrease its height by a lot.}\uk{agree maybe only table here or to make something prettier? }} 
    \label{Figures:model_scales}
\end{figure}
}

\begin{table}
    \setlength\tabcolsep{5.0pt}
    \centering\footnotesize
    \begin{tabular}{lllr}
         \textbf{Parameters}&\textbf{Concept} & \textbf{Concept}  & \textbf{Relation}  \\
         &\textbf{Recall} & \textbf{Precision} & \textbf{Coverage} \\ \toprule
         350M  & 0.83$\pm$0.01 & 0.89$\pm$0.01 & 0.21$\pm$0.01   \\ 
         1.3B & 0.93$\pm$0.01 & 0.84$\pm$0.01 & 0.37$\pm$0.01  \\ 
         6.7B  & 0.92$\pm$0.01 & 0.83$\pm$0.02 & 0.42$\pm$0.02   \\
         175B & 0.97$\pm$0.0 & 0.88$\pm$0.0 & 0.43$\pm$0.03   \\
         \bottomrule
    \end{tabular}
    \caption{Model size and performance comparison on the development set of \stqa{}. The relation coverage improves as model size is increased. }
    \label{table:implicit_relation_model_scale}
\end{table}

\section{Implicit Relations for QA}
\label{sec:qa_performance}

Given that LMs infer implicit relations well, a natural question is whether they improve performance on answering implicit reasoning questions.  

\begin{figure}[t]
    \centering
    \includegraphics[scale=0.5]{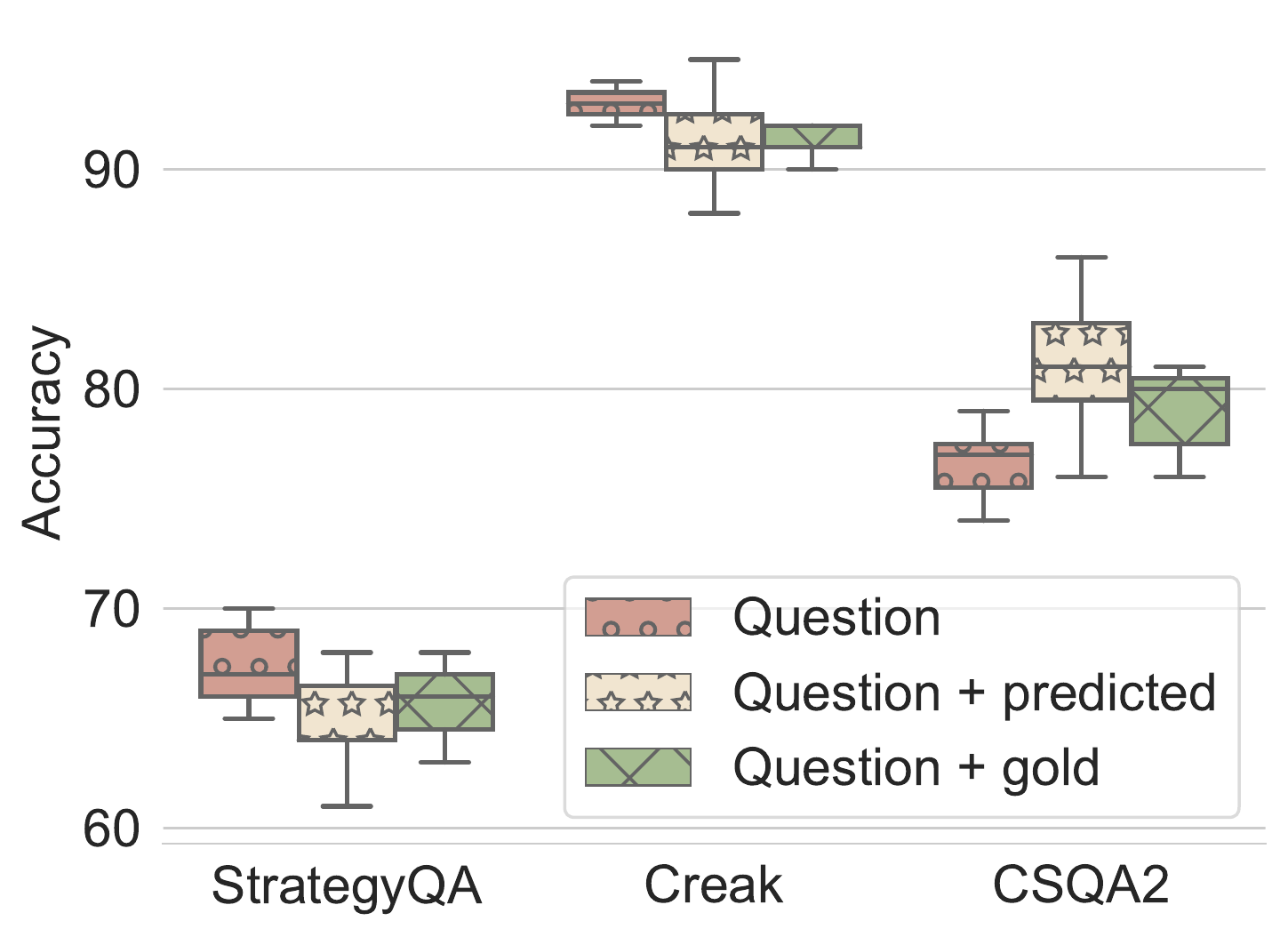}
    \caption{Test QA accuracy under all conditions, averaged over 7 seeds. Providing the gold implicit relations did not contribute to model performance.} 
    \label{Figures:QA}
\end{figure}

To examine this, we created three experimental setups: \textbf{\textit{Question + predicted}}: in-context examples are triples of the question, the implicit relations, and the True/False answer; the model is given a question and asked to return the implicit relations and the answer. \textbf{\textit{Question + gold}}: Similar to \textit{Question + predicted} except that the model is given the target question and gold implicit relations and asked to return the answer. \textbf{\textit{Question only}}: in-context examples are pairs of questions and answers, and the model is given a question and asked to provide the answer. We report an average over 7 seeds. See Figure~\ref{Figures:QA} for results.  

%We observed that the \textit{Question} performance for \strategy{} in our experiments is similar to accuracy scores reported by \cite{bigbench}.

Overall, access to either gold or predicted relations does not improve accuracy. This suggests that additional capabilities are missing from LMs to handle implicit reasoning questions, such as retrieval and reasoning. 
This agrees with work on chain-of-thoughts prompting \cite{Wei2022ChainOT}, which found that adding an explanation of the reasoning process to a question does not improve performance on \stqa{} for both GPT-3 and LaMDA \cite{Thoppilan2022LaMDALM}. Nevertheless, recently \newcite{chowdhery2022palm} achieved improvements on \stqa{} using chain-of-thought prompting, but with the larger 540B-parameter PaLM.

%Recently, \cite{Wei2022ChainOT} showed that adding chain of thought, a kind of reasoning decomposition, to the prompt of GPT-3 and LaMDA 137B did not manage to improve QA task for \strategy{}, but when they used chain of thought prompting with PaLM 540B parameters model \cite{chowdhery2022palm}, they achieved strong performance improvement relative to their baselines. Another recent paper \cite{Lampinen2022CanLM} showed that the benefit of large LM from augmenting reasoning and explanation in in-context learning is an emergent capability and related closely to the model size. 

On \stqa{}, adding gold relations (\textit{Question + gold}) does not improve QA performance.\footnote{paired t-test with $p$-value $> 0.05$.} Additionally, no significant differences were observed when the model inferred implicit relations on its own (\textit{Question + predicted}).\footnote{paired t-test with $p$-value $> 0.05$.} 
For \creak{}, adding gold implicit relations did not improve accuracy compared to \textit{Question only}, and none of the experiments showed any significant difference. Last, in \csqa{} adding gold implicit relations (\textit{Question + gold}) did not improve the QA performance, but we observed a statistically significant accuracy improvement of 4.5\% when the model inferred the implicit relations (\textit{Question + predicted}).\footnote{paired t-test with $p$-value $< 0.05$, Cohen's $d = 1.7$.}

To further analyze the results, we computed the point-biserial correlation coefficient between the relation coverage score and the binary outcome (correct/incorrect) for each question. We found that relation coverage score and answer accuracy are entirely not correlated with a $r_{\text{pb}}$ coefficient of $0.03$, $-0.02$ and $0.06$ for \stqa{}, \csqa{} and \creak{} respectively. Overall, our results indicate that inferring implicit relations correctly is not sufficient to answer implicit reasoning questions.
\section{Related Work}
\label{sec:related_work}

Recent work utilized the ability of large LMs to generate intermediate reasoning steps for improving performance on QA tasks \cite{Wei2022ChainOT,wang2022self, zelikman2022star,nye2022show}. 
\citet{Wei2022ChainOT} introduced `chain-of-thought' prompting to elicit intermediate reasoning steps along with answers from LMs, 
% \cite{Wei2022ChainOT} introduced chain-of-thought prompting as a method for inducing a series of short sentences that mimic the reasoning process. Using this method, models managed to elicit chain-of-thought together with answers, 
which improved performance on several reasoning tasks. 
% Other papers have suggested more methods to improve downstream tasks with reasoning generation of intermediate steps \cite{zelikman2022star,nye2022show}.
Conversely, we propose a task and benchmark for evaluating the ability of LMs to infer the intermediate reasoning steps themselves.

% Prior work has dealt with externalizing reasoning abilities in LMs. \newcite{Gu2021DREAMUM} trained models to elaborate on situational questions. \newcite{shwartz-etal-2020-unsupervised} used ``self-talk'' to generate clarification questions that uncover additional background knowledge for commonsense reasoning questions. \newcite{NEURIPS2020_e992111e} trained a model to reason implicit knowledge. We contribute to this effort by defining implicit relations, and evaluating the in-context ability to infer them with large LMs.
Prior work has dealt with reasoning abilities in LMs \cite{NEURIPS2020_e992111e,khashabi-etal-2020-unifiedqa,Gu2021DREAMUM} by fine-tuning LMs to generate additional knowledge for reasoning questions. We contribute to this effort by evaluating the in-context ability to infer implicit relations with large LMs.

% JB: bring back from camera ready
Implicit relations are closely related to question decomposition, which have been used in past work to improve performance on questions that require reasoning \cite{min-etal-2019-multi,wolfson2020break,perez-etal-2020-unsupervised,khot-etal-2021-text}.  We contribute to this research direction by defining implicit relations pairs, which provide a structured representation of the decomposed sub-questions and allow us to examine how language models infer reasoning steps. Several works in narrative understanding \cite{rajpurkar-etal-2018-know,mostafazadeh-etal-2020-glucose,lal-etal-2021-tellmewhy} have attempted to assess a model's implicit reasoning capabilities using different methods, such as assessing the solution path to unanswerable questions and and narrative understanding through question-answering. Despite their different approaches, these studies are relevant to our cause. 

%\uk{*} 
% \paragraph{Missing elements}
%Implicit relations also relate to the range of natural language phenomena that involve ``missing elements'', i.e., elements that are not explicitly expressed in the text, but can still be reliably inferred by the reader \cite{rosiger2018bridging,cheng-erk-2018-implicit,elazar-goldberg-2019-wheres,hou-2020-bridging,tne,pyatkin-etal-2020-qadiscourse,anthonio-roth-2021-resolving}.
\section{Conclusion}
\label{sec:conclusion}

We propose the task of implicit relation inference, which decouples inference of reasoning steps from their execution. We introduce \imprel{}, a benchmark that includes more than 2,000 annotated implicit relations.
We show large LMs can infer implicit relations across multiple types of questions and reasoning skills, but this success does not translate to an improvement in answering implicit reasoning questions. Our work sheds light on capabilities missing from large LMs for addressing implicit reasoning questions, and provides a valuable resource for improving the ability of models to infer implicit relations.

\section*{Limitations}
This research has some limitations, which are typical for work on text generation with large language models.

First, we demonstrated that large LMs can infer implicit relations from complex questions, but we also showed that they may fail to answer those questions correctly. It is unclear how LMs can use implicit relations to improve QA accuracy or what is the path that leads from inferring implicit relations to actually answering the questions. 

Second, evaluating relation coverage
requires comparing to free texts, and therefore may be prone to error. Despite the fact that an analysis performed manually exhibited a high degree of consistency with the automatic one, we cannot guarantee the same result for datasets or parameters that have not been tested.

Finally, our research was conducted utilizing OpenAI's GPT-3 family of models, which are not publicly available. Despite our best efforts to eliminate confounding factors, there is a lack of transparency regarding the training methods and the data composition used for pretraining those models.

% \section*{Ethics Statement}
% Scientific work published at EMNLP 2022 must comply with the \href{https://www.aclweb.org/portal/content/acl-code-ethics}{ACL Ethics Policy}. We encourage all authors to include an explicit ethics statement on the broader impact of the work, or other ethical considerations after the conclusion but before the references. The ethics statement will not count toward the page limit (8 pages for long, 4 pages for short papers).

\section*{Acknowledgements}
We thank Itay Levy for useful feedback. This research was partially supported by the Computer Science Scholarship granted by the Séphora Berrebi Foundation, the Yandex Initiative for Machine Learning, and the European Research Council (ERC) under the European
Union Horizons 2020 research and innovation programme (grant ERC DELPHI 802800).

\bibliography{anthology,custom}
\bibliographystyle{acl_natbib}

\appendix

\appendix
\section{Examples and Qualitative Analysis}
\label{sec:example_appendix}
\subsection{Qualitative analysis of evaluation}
In addition to the evaluation described in \S\ref{sec:experimental_setting} and used throughout our study, we performed manual qualitative analysis to assess the relation coverage metric model outputs.
We sampled 50 examples randomly from \stqa{} together with \gpt{} predictions and manually labeled if the implicit relations output by the model are semantically correct.

\begin{table}[t]
    \setlength\tabcolsep{3.5pt}
    \centering\footnotesize
    \begin{tabular}{lc}
         & Relation coverage   \\ \midrule
         Manual evaluation  & 0.6  \\
         Automatic evaluation  & 0.53   \\
         \bottomrule
    \end{tabular}
    \caption{Manual and automatic evaluation of 50 examples from \stqa{}.}
    \label{table:manual_analysis}
\end{table}

Results in Table~\ref{table:manual_analysis} show that our relation coverage was slightly more conservative than our manual evaluation -- 0.53 vs. 0.6.
Out of the 20 examples that we marked as incorrect, the automatic evaluation scored 13 examples (65\%) with 0 relation coverage, 6 (30\%) examples with partial coverage (scored 0.5), which indeed we included a partially correct prediction but not enough to to cover the needed reasoning process. Only one example out of the 20 was a false positive.

\subsection{\imprel{} Examples}
Tables~\ref{table:strqa_examples},\ref{table:creak_examples},\ref{table:csqa_examples} present examples from each source dataset accompanied by the question, gold annotated implicit relation pairs, predicted pairs generated by the model and the answer to the question.
\begin{table*}[ht]
    \centering
    \footnotesize
    \begin{tabular}{p{14.6cm}}
         Examples from \strategy{}\\ \toprule
         \textbf{Q: }\textit{Could Eric Clapton's children play a regulation game of basketball among themselves?} 
         \newline \textbf{G: }\pair{Eric Clapton}{number of children} , \pair{regulation game of basketball}{number of players}
         \newline \textbf{P: }\pair{eric clapton}{number of children} , \pair{regulation game of basketball}{number of players}
         \newline \textbf{A: }\texttt{False}\\\midrule

         \textbf{Q: }\textit{Can you purchase General Motors products at a movie theater?}
         \newline \textbf{G: }\pair{General Motors}{Types of products} , \pair{Movie theater}{products sold}
         \newline \textbf{P: }\pair{general motors}{type of product} , \pair{movie theater}{type of product}
         \newline
         \textbf{A: }\texttt{False}\\ \midrule
         
         \textbf{Q: }\textit{Does Rupert Murdoch's alma mater have more history than the USA?}
         \newline \textbf{G: }\pair{Rupert Murdoch's alma mater}{year founded} , \pair{USA}{year founded}
         \newline \textbf{P: }\pair{rupert murdoch's alma mater}{date of founding} , \pair{usa}{date of founding}
         \newline
         \textbf{A: }\texttt{True}\\ \midrule

         \textbf{Q: }\textit{Do Squidward Tentacles and Alan Greenspan have different musical passions?}
         \newline \textbf{G: }\pair{Squidward Tentacles}{instruments played} , \pair{Alan Greenspan}{instruments played}
         \newline \textbf{P: }\pair{squidward tentacles}{musical instrument} , \pair{alan greenspan}{musical instrument}
         \newline
         \textbf{A: }\texttt{False}\\ \midrule
         
         \textbf{Q: }\textit{Is the most expensive color in the world Blue?}
         \newline \textbf{G: }\pair{Blue}{cost of pigment} , \pair{most expensive color in the world}{cost of pigment}
         \newline \textbf{P: }\pair{blue}{color} , \pair{most expensive color}{color}
         \newline
         \textbf{A: }\texttt{True}\\ \bottomrule
         
     \end{tabular}
    \caption{Examples of gold annotations (\textbf{G}) and predicted (\textbf{P}) concept-relation pairs from \imprel{} along with the question (\textbf{Q}) and answer (\textbf{A}) from \strategy{}.}
    \label{table:strqa_examples}
\end{table*}

\begin{table*}[ht]
    \centering
    \footnotesize
    \begin{tabular}{p{14.6cm}}
         Examples from \creak{}\\ \toprule
         \textbf{Q: }\textit{People celebrate Cinco de Mayo before summer starts.}
         \newline \textbf{G: }\pair{Cinco de Mayo}{date} , \pair{summer}{starting date}
         \newline \textbf{P: }\pair{cinco de Mayo}{date}
         \newline \textbf{A: }\texttt{True}\\\midrule

         \textbf{Q: }\textit{Arthur C. Clarke wrote a book set in outer space.}
         \newline \textbf{G: }\pair{Arthur C. Clarke}{writing genre}
         \newline \textbf{P: }\pair{arthur c. clarke}{book genre}
         \newline
         \textbf{A: }\texttt{True}\\ \midrule
         
         \textbf{Q: }\textit{The local chamber music company hired a Harp seal to join the group.}
         \newline \textbf{G: }\pair{Harp seal}{musical ability}
         \newline \textbf{P: }\pair{harp seal}{natural habitat}
         \newline
         \textbf{A: }\texttt{False}\\ \midrule

         \textbf{Q: }\textit{Nicole Kidman has never heard of Batman.}
         \newline \textbf{G: }\pair{Nicole Kidman}{filmography}
         \newline \textbf{P: }\pair{nicole kidman}{knowledge}
         \newline
         \textbf{A: }\texttt{False}\\ \midrule
         
         \textbf{Q: }\textit{Competitors know how to drive when they are competing in Formula One racing.}
         \newline \textbf{G: }\pair{Formula One racing}{type of sport}
         \newline \textbf{P: }\pair{formula one racing}{skills required}
         \newline
         \textbf{A: }\texttt{True}\\ \bottomrule
         
     \end{tabular}
    \caption{Examples of gold annotations (\textbf{G}) and predicted (\textbf{P}) concept-relation pairs from \imprel{} along with the question (\textbf{Q}) and answer (\textbf{A}) from \creak{}.}
    \label{table:creak_examples}
\end{table*}

\begin{table*}[ht]
    \centering
    \footnotesize
    \begin{tabular}{p{14.6cm}}
         Examples from \csqa{}\\ \toprule
         \textbf{Q: }\textit{Trees are never part of forests.}
         \newline \textbf{G: }\pair{forests}{type of flora}
         \newline \textbf{P: }\pair{trees}{location}
         \newline \textbf{A: }\texttt{False}\\\midrule

         \textbf{Q: }\textit{You can only win a medal at the Olympics if you place in the top 3?}
         \newline \textbf{G: }\pair{Olympics}{medal awarding rules}
         \newline \textbf{P: }\pair{medal}{requirements for winning}
         \newline
         \textbf{A: }\texttt{True}\\ \midrule
         
         \textbf{Q: }\textit{Is it true that a hospital is capable of treating Dutch Elm disease?}
         \newline \textbf{G: }\pair{Dutch Elm disease}{targeted organism} , \pair{hospital}{species treated}
         \newline \textbf{P: }\pair{hospital}{type of diseases treated}
         \newline
         \textbf{A: }\texttt{False}\\ \midrule

         \textbf{Q: }\textit{at least one letter in the word walnut is q.}
         \newline \textbf{G: }\pair{walnut}{spelling}
         \newline \textbf{P: }\pair{word walnut}{letters}
         \newline
         \textbf{A: }\texttt{False}\\ \midrule
         
         \textbf{Q: }\textit{Is the temperature usually quite hot before Christmas in the southern hemisphere?}
         \newline \textbf{G: }\pair{Christmas}{month} , \pair{southern hemisphere}{summer months}
         \newline \textbf{P: }\pair{southern hemisphere}{seasons}
         \newline
         \textbf{A: }\texttt{True}\\ \bottomrule
         
     \end{tabular}
    \caption{Examples of gold annotations (\textbf{G}) and predicted (\textbf{P}) concept-relation pairs from \imprel{} along with the question (\textbf{Q}) and answer (\textbf{A}) from \csqa{}.}
    \label{table:csqa_examples}
\end{table*}

\section{Number of Prompt Examples}
\label{sec:prompt_size_appendix}
We investigate how different number of examples influence implicit relation inference. We run the original experiment with $k \in \{8, 16, 32\}$ examples in the prompt for 3 random seeds. The results (Table~\ref{table:prompt_size}) show that the number of examples in the prompt has little effect on all evaluation metrics, justifying our choice to use $k=16$ in all experiments.

\begin{table}
    
    \centering\footnotesize
    % \begin{tabular}{l|c|c|c|c}
    \begin{tabular}{llllr}
         \toprule
         &\textbf{$k$}&\textbf{Concept} & \textbf{Concept}  & \textbf{Relation}  \\
         &&\textbf{Recall} & \textbf{Precision} & \textbf{Coverage}  \\ \midrule
         
         \multirow{3}{*}{\stqa{}} & 8  & 0.94$\pm$0.02 & 0.87$\pm$0.01 & 0.48$\pm$0.02   \\ 
         &16& 0.97$\pm$0.01 & 0.88$\pm$0.01 & 0.53$\pm$0.02   \\
         &32& 0.97$\pm$0.01 & 0.88$\pm$0.01 & 0.50$\pm$0.01\\\cmidrule{2-5}
         
         \multirow{3}{*}{\creak{}}  & 8  & 0.94$\pm$0.02 & 0.91$\pm$0.02 & 0.56$\pm$0.0   \\ 
         &16& 0.93$\pm$0.02 & 0.91$\pm$0.01 & 0.59$\pm$0.03   \\
         &32& 0.95$\pm$0.01 & 0.90$\pm$0.02 & 0.62$\pm$0.02\\\cmidrule{2-5}
         
         \multirow{3}{*}{\csqa{}}  & 8  & 0.94$\pm$0.0 & 0.95$\pm$0.02 & 0.61$\pm$0.02   \\ 
         &16& 0.95$\pm$0.01 & 0.96$\pm$0.0 & 0.67$\pm$0.01   \\
         &32& 0.95$\pm$0.01 & 0.96$\pm$0.02 & 0.66$\pm$0.02\\\bottomrule

    \end{tabular}
    \caption{Development set results for 8/16/32 examples in the prompt, averaged for 3 seeds.} 
    \label{table:prompt_size}
\end{table}

\section{Annotation Task Instruction}
\label{sec:mturk}
\subsection{Task instruction}
\includepdf[scale=0.75]{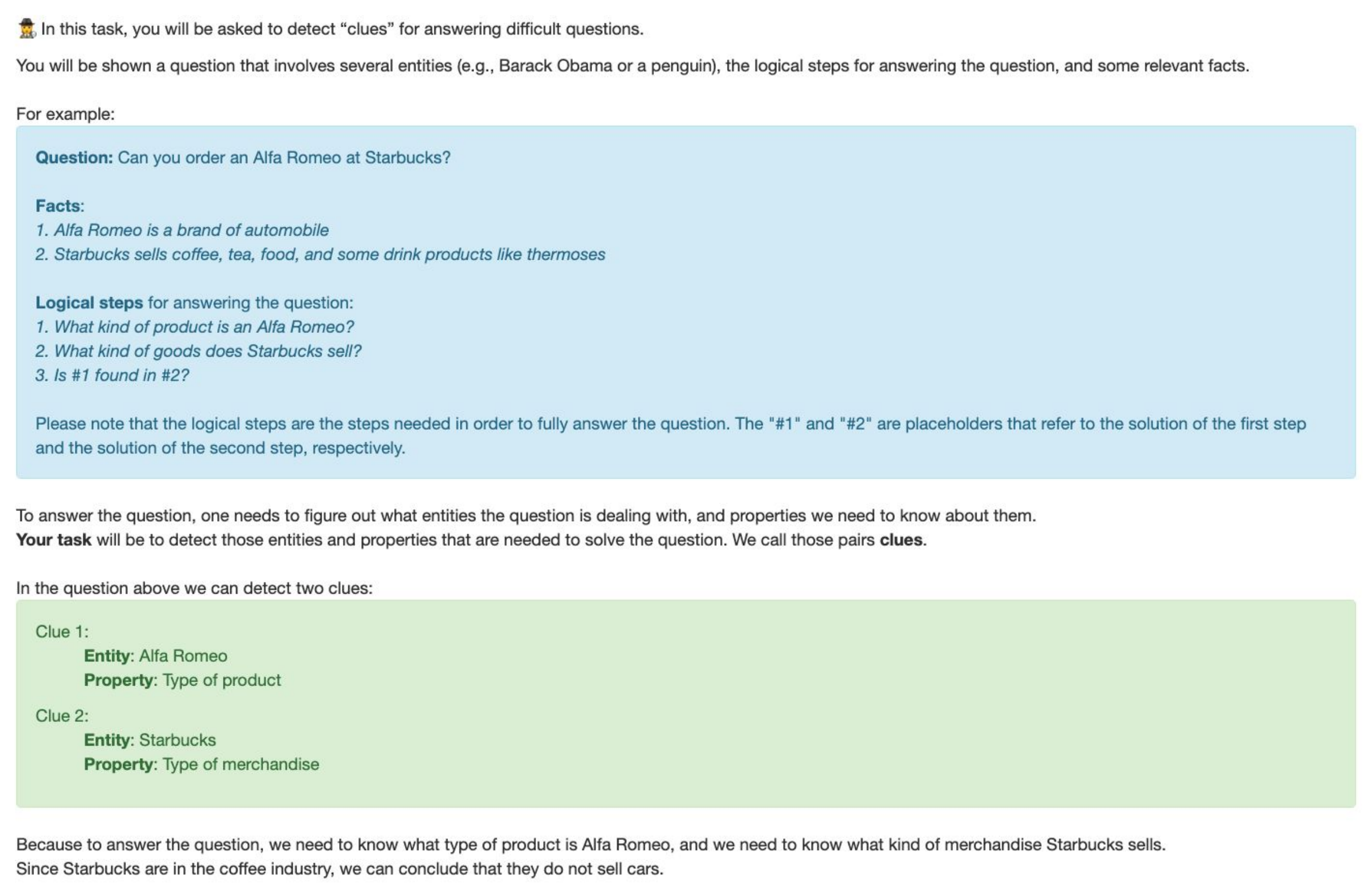}
\subsection{Task instruction cont.}
\includepdf[scale=0.75]{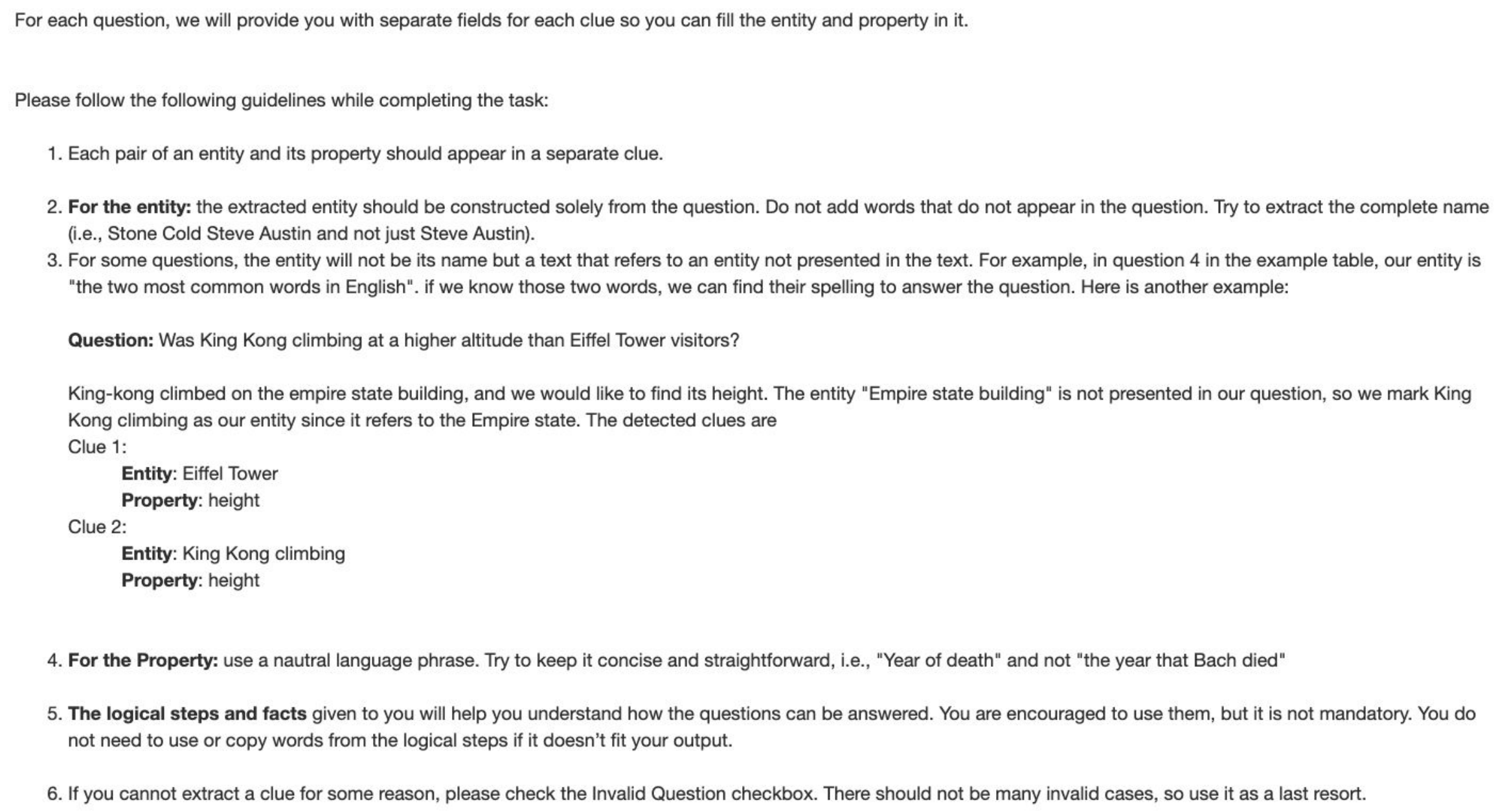}
\subsection{Concept-relation pairs}


\section*{Supplementary material}

# C  Annotation Task Instruction

## C.1  Task instruction

🕵️ In this task, you will be asked to detect "clues" for answering difficult questions.

You will be shown a question that involves several entities (e.g., Barack Obama or a penguin), the logical steps for answering the question, and some relevant facts.

For example:

> **Question:** Can you order an Alfa Romeo at Starbucks?
>
> **Facts**:
> 1. Alfa Romeo is a brand of automobile
> 2. Starbucks sells coffee, tea, food, and some drink products like thermoses
>
> **Logical steps** for answering the question:
> 1. What kind of product is an Alfa Romeo?
> 2. What kind of goods does Starbucks sell?
> 3. Is #1 found in #2?
>
> Please note that the logical steps are the steps needed in order to fully answer the question. The "#1" and "#2" are placeholders that refer to the solution of the first step and the solution of the second step, respectively.

To answer the question, one needs to figure out what entities the question is dealing with, and properties we need to know about them.
**Your task** will be to detect those entities and properties that are needed to solve the question. We call those pairs **clues**.

In the question above we can detect two clues:

> Clue 1:
>   **Entity**: Alfa Romeo
>   **Property**: Type of product
>
> Clue 2:
>   **Entity**: Starbucks
>   **Property**: Type of merchandise

Because to answer the question, we need to know what type of product is Alfa Romeo, and we need to know what kind of merchandise Starbucks sells.
Since Starbucks are in the coffee industry, we can conclude that they do not sell cars.

## C.2 Task instruction cont.

For each question, we will provide you with separate fields for each clue so you can fill the entity and property in it.

Please follow the following guidelines while completing the task:

1. Each pair of an entity and its property should appear in a separate clue.

2. **For the entity:** the extracted entity should be constructed solely from the question. Do not add words that do not appear in the question. Try to extract the complete name (i.e., Stone Cold Steve Austin and not just Steve Austin).

3. For some questions, the entity will not be its name but a text that refers to an entity not presented in the text. For example, in question 4 in the example table, our entity is "the two most common words in English". if we know those two words, we can find their spelling to answer the question. Here is another example:

   **Question:** Was King Kong climbing at a higher altitude than Eiffel Tower visitors?

   King-kong climbed on the empire state building, and we would like to find its height. The entity "Empire state building" is not presented in our question, so we mark King Kong climbing as our entity since it refers to the Empire state. The detected clues are
   Clue 1:
       **Entity**: Eiffel Tower
       **Property**: height
   Clue 2:
       **Entity**: King Kong climbing
       **Property**: height

4. **For the Property:** use a nautral language phrase. Try to keep it concise and straightforward, i.e., "Year of death" and not "the year that Bach died"

5. **The logical steps and facts** given to you will help you understand how the questions can be answered. You are encouraged to use them, but it is not mandatory. You do not need to use or copy words from the logical steps if it doesn't fit your output.

6. If you cannot extract a clue for some reason, please check the Invalid Question checkbox. There should not be many invalid cases, so use it as a last resort.

## C.3 Concept-relation pairs

**Question:** ${question}

**Logical Steps:** ${decomposition}

**Facts:** ${facts}

**Clues:** (at least one)

Clue 1:
- Entity
  - [ Write the first Entity here... ]
- Property
  - [ Write the first Property here... ]

Clue 2:
- Entity
  - [ Write Entity here... ]
- Property
  - [ Write Property here... ]



\label{sec:appendix}
\end{document}